\pdfoutput=1
\documentclass[conference]{IEEEtran}
\usepackage{amsmath,amssymb,amsfonts,mathrsfs,bm}
\usepackage{amstext}
\usepackage{upgreek}
\usepackage{multicol}
\usepackage{indentfirst}
\usepackage{graphicx}
\usepackage{paralist}
\usepackage{multirow}
\usepackage{tabularx}
\usepackage[noadjust]{cite}

\usepackage{booktabs}
\usepackage{subfigure}
\usepackage{color}
\usepackage{psfrag}

\IEEEoverridecommandlockouts
\allowdisplaybreaks
\usepackage{algorithm,algorithmic}
\floatname{algorithm}{\small \bf Algorithm}

\newtheorem{remark}{\bf Remark}

\usepackage{accents}

\allowdisplaybreaks
\begin{document}
\title{Model-Free Unsupervised Learning for Optimization Problems with Constraints}
\author{
 	\IEEEauthorblockN{Chengjian Sun, Dong Liu, and Chenyang Yang}\\
	\IEEEauthorblockA{Beihang University, Beijing, China\\
	Email: \{sunchengjian, dliu, cyyang\}@buaa.edu.cn}
}
\maketitle

\begin{abstract}
In many optimization problems in wireless communications, the expressions of objective function or constraints are hard or even impossible to derive, which makes the solutions difficult to find. In this paper, we propose a model-free learning framework to solve constrained optimization problems without the supervision of the optimal solution. Neural networks are used respectively for parameterizing the function to be optimized, parameterizing the Lagrange multiplier associated with instantaneous constraints, and approximating the unknown objective function or constraints. We provide learning algorithms to train all the neural networks simultaneously, and reveal the connections of the proposed framework with reinforcement learning. Numerical and simulation results validate the proposed framework and demonstrate the efficiency of model-free learning by taking power control problem as an example.
\end{abstract}

\section{Introduction}
Various resource allocation and transceivers in wireless networks, such as power allocation, beamforming, and caching policy, can be designed by solving optimization problems with constraints, say imposed by the maximal transmit power, cache size, and the minimal data rate requirement~\cite{kong2018hybrid,liu2018caching}.

Depending on the applications, the objective function, constraints and the policy to be optimized may vary in different timescales.
If they are in the same timescale, the problem is variable optimization, and the policy to be optimized is a vector with finite dimension, e.g., optimizing beamforming based on small scale channel gains to maximize the instantaneous data rate subject to instantaneous transmit power constraint. If they are in different timescales, the problem is functional optimization~\cite{Zeidler2013Functional}, i.e., the policy to be optimized is a function, which can be interpreted as a vector with infinite elements. A classical example of functional optimization is finding the instantaneous power allocation to maximize the ergodic capacity under the average power constraint, whose solution is the classical water-filling power allocation~\cite{goldsmith2005wireless}.

Variable optimizations have been well studied. Efficient tools, such as interior point method~\cite{boyd2004convex}, have been developed to find the numerical optimal solutions of convex optimizations and various approximation methods have been widely used in non-convex optimizations. The optimal solutions of functional optimizations are, however, generally not in closed-form and inefficient to be obtained numerically. One way for numerical searching is the finite element method~\cite{Zienkiewicz1977FEM}, which converts the functional optimization into a variable optimization by only optimizing the values of function on a finite sampled points. However, such method suffers from the curse of dimensionality. To overcome this shortage, an unsupervised learning framework was developed in~\cite{Chengjian2019GC}, which parameterizes the functions by neural networks and trains the network parameters with stochastic gradient descent (SGD).

To apply numerical searching methods, the expressions of objective function and constraints, i.e., the model, should be known. For methods like interior point method and SGD, the gradients of objective function and constraints with respect to the optimization variables or functions are further required. However, in many scenarios, the expressions of the objective or constraints are unavailable, or too complex to derive their gradients.
Finite difference method can be used to estimate the gradients according to the observations of the objective and constraints. However, such a method is inefficient when the objective function and the constraints are in high dimensions, and hence is not applicable for functional optimization problems, which are with infinite dimensions.

In this work, we propose a model-free framework to solve functional optimizations with instantaneous and average constraints without the supervision of optimal solution.
We begin with a model-based framework for unsupervised learning where a neural network (called as \emph{policy network}) is used for parameterizing the function to be optimized. We recast the original constrained problem in the dual domain where the Lagrangian is served as the objective function and another neural network (called as \emph{multiplier network}) is introduced to parameterize the Lagrange multiplier associated with the instantaneous constraint.
In the model-free framework, we resort to neural networks (called as \emph{value networks}) to approximate the unavailable expressions of objective function or constraints so that their gradients can be obtained for training the policy and multiplier networks.
Then, we reveal the connections of the proposed framework with reinforcement learning and show how to extend the proposed framework into stochastic policy optimization, which is applicable for both continuous and discrete policy optimizations. We study a simple power control problem
to illustrate how to apply the framework and show the effectiveness of the model-free unsupervised learning.

\section{Unsupervised Learning for Optimizations}
In this section, we introduce model-based and model-free unsupervised learning frameworks for solving general functional optimization problems. Since variable optimization can be treated as a special case of functional optimization, these frameworks are also applicable to variable optimization.

Let $\mathbf h  \in\mathbb{R}^n$ denote a random vector reflecting environment status, e.g., channel gains. For each realization of $\mathbf{h}$, we aim to find a vector $\mathbf x \in \mathbb{R}^{m}$ to execute, e.g., transmit powers. Let function (called as a policy) $\mathbf f$: $\mathbb{R}^n \mapsto \mathbb{R}^m$  denote the mapping from $\mathbf{h}$ to $\mathbf x$, i.e., $\mathbf{x} = \mathbf{f} (\mathbf{h})$. The performance metric is a scalar function of $\mathbf x$ and $\mathbf h$ denoted by $J(\mathbf{x}, \mathbf{h})$, e.g., the instantaneous data rate. The goal is to design a function~$\mathbf f$ that maximizes the performance metric averaged over $\mathbf h$, i.e., $\mathbb{E}_{\mathbf h} [J(\mathbf{x}, \mathbf{h})]$, subject to some constraints. This can be formulated as a general functional optimization problem as follows,
\begin{subequations} \label{eqn:p1}
	\begin{align}
  \hspace{-1cm}\mathsf{P1}:\quad	\max_{\mathbf f(\mathbf h)} ~& \mathbb{E}_{\mathbf h} [ J(\mathbf f(\mathbf h), \mathbf{h})] \label{eqn:obj} \\
	\rm{s.t.} ~ & \mathbf g(\mathbf f(\mathbf h), \mathbf{h}) \preceq \mathbf{0} \label{eqn:short} \\
	& \mathbb{E}_{\mathbf h} [ \mathbf c(\mathbf f(\mathbf h), \mathbf{h}) ]\preceq \mathbf{0} \label{eqn:long}
	\end{align}
\end{subequations}
where~\eqref{eqn:short} and~\eqref{eqn:long} denote the instantaneous and the average constraints, respectively, and the curled inequality symbol ``$\preceq$" (or ``$\succeq$") denotes the element-wise inequality. It is noteworthy that optimization problems minimizing the objective function or having ``$\succeq$" or ``$=$" constraints (e.g., minimal data rate constraint) can be easily transformed into problem $\mathsf{P1}$.

In general, problem $\mathsf{P1}$ is hard to solve because it is a function $\mathbf{f}(\mathbf{h})$ that needs to optimize, which can be interpreted as vectors with infinite dimension when $\mathbf h$ is with infinity number of possible values.



\subsection{Model-Based Unsupervised Learning}

To tackle with the constraints, we reconsider problem~$\mathsf{P1}$ in its dual domain. The Lagrangian of $\mathsf{P1}$ can be written as~\cite{gregory2018constrained}
\begin{align}
&\mathcal{L}(\mathbf f(\mathbf{h}), \bm \lambda(\mathbf h), \bm \xi) \nonumber \\
&= \mathbb{E}_{\mathbf h} \left[ J(\mathbf f(\mathbf h), \mathbf{h}) - \bm \lambda(\mathbf h)^T \mathbf{g}(\mathbf f(\mathbf h), \mathbf{h}) - \bm\xi^T \mathbf{c}(\mathbf f(\mathbf h), \mathbf h)  \right] \label{eqn:lag}
\end{align}
where $\bm \lambda(\mathbf h)$ and $\bm \xi$ are the Lagrange multiplier associated with constraints~\eqref{eqn:short} and~\eqref{eqn:long}, respectively. When strong duality condition holds~\cite{boyd2004convex}, the original problem is equivalent to finding the saddle point of the Lagrangian as
\begin{subequations}
	\begin{align}
	\hspace{-1cm}\mathsf{P2}:\quad	\min_{\bm \lambda(\mathbf h), \bm \xi} \max_{\mathbf f(\mathbf h)}  ~& \mathcal{L}(\mathbf f(\mathbf{h}), \bm \lambda(\mathbf h), \bm \xi) \\
	\rm{s.t.} ~~&  \bm \lambda(\mathbf h) \succeq \mathbf{0} \\
	&  \bm \xi \succeq \mathbf{0}
	\end{align}
\end{subequations}
where $\bm \lambda({\mathbf h})$ is also a function that needs to be optimized. Thanks to the universal approximation theorem~\cite{Hornik1989UnivApprox}, we can introduce two neural networks to approximate function $\mathbf{f}$ as $\mathbf{f}(\mathbf h) \approx \tilde{\mathbf{f}}(\mathbf h;\bm \theta_f)$ (called as \emph{policy network}) and approximate multiplier as $\bm \lambda(\mathbf h) \approx \tilde{\bm \lambda}(\mathbf h; \bm \theta_\lambda)$ (called as \emph{multiplier network}), respectively, with arbitrary accuracy by finite-dimension parameter vectors $\bm \theta_f$ and $\bm \theta_\lambda$. Then, problem $\mathsf{P2}$ degenerates into the following variable optimization as
\begin{subequations}
	\begin{align}
	\hspace{-1cm}\mathsf{P3}:\quad	 \min_{\bm \theta_\lambda, \bm \xi} \max_{\bm \theta_{f}}~&  \mathcal{L}( \tilde{\mathbf f}(\mathbf{h};\bm \theta_f), \tilde{\bm \lambda}(\mathbf h; \bm \theta_\lambda), \bm \xi) \label{eqn:Lag}  \\
	\rm{s.t.} ~~& \tilde{\bm \lambda}(\mathbf h, \bm \theta_\lambda) \succeq \mathbf{0} \label{eqn:g}\\
	& \bm \xi \succeq \mathbf{0}  \label{eqn:c}
	\end{align}
\end{subequations}

To solve problem $\mathsf{P3}$, we can adopt the primal-dual stochastic gradient method~\cite{Chengjian2019PIMRC} that iteratively updates the primal variable $\bm \theta_f$, and the dual variables $\bm \theta_\lambda$ and  $\bm \xi$ along the ascent and descent directions of sample-averaged gradients, respectively. The gradients of the Lagrangian~\eqref{eqn:Lag} with respect to $\bm \theta_f$, $\bm \theta_\lambda$, and $\bm \xi$ can be derived as
\begin{subequations} \label{eqn:grad}
\begin{align}
\nabla_{\bm \theta_f} \mathcal{L} =  &\mathbb{E}_{\mathbf h}\Big[ \nabla_{\bm \theta_f} \tilde{\mathbf{f}}(\mathbf h; \bm \theta_{f})\big[\nabla_{{\mathbf x}}  J(\mathbf x, \mathbf h)  - \nabla_{{\mathbf x}} \mathbf{g}(\mathbf x, \mathbf h) \tilde{\bm \lambda}(\mathbf{h};\bm \theta_\lambda ) \nonumber \\
& - \nabla_{\mathbf{x}} \mathbf c(\mathbf x, \mathbf h) \bm \xi\big] \big|_{\mathbf x = \tilde{\mathbf{f}}(\mathbf h; \bm \theta_{f}) }  \Big] \label{eqn:pg} \\
\nabla_{\bm \theta_\lambda}  \mathcal{L} = &  -\mathbb{E}_{\mathbf h} \left[  \nabla_{\bm \theta_\lambda} \tilde{\bm \lambda} (\mathbf h; \bm \theta_\lambda) \mathbf{g}(\tilde{\mathbf{f}}(\mathbf h; \bm \theta_{f}),\mathbf h) \right] \\
\nabla_{\bm \xi}  \mathcal{L} = &-\mathbb{E}_{\mathbf h} \left[\mathbf c(\tilde{\mathbf f}(\mathbf h; \bm \theta_f), \mathbf{h})\right]
\end{align}
\end{subequations}
where $\nabla_{\mathbf x}  y  = [\frac{\partial y}{\partial x_1}, \cdots, \frac{\partial y}{\partial x_n} ]^T$ denotes the gradient, $\nabla_{\mathbf x} \mathbf y = [(\nabla_{\mathbf x}  y_1), \cdots, (\nabla_{\mathbf x}  y_m)]$ denotes the transpose of Jacobian matrix and $(\cdot)^T$ is the transpose operation. Both $ \nabla_{\bm \theta_f} \tilde{\mathbf{f}}(\mathbf h; \bm \theta_{f})$ and $\nabla_{\bm \theta_\lambda} \tilde{\bm \lambda} (\mathbf h; \bm \theta_\lambda)$ can be computed via back propagation.

Let $\mathcal{B}$ denote a batch of realizations of $\mathbf h$. Then, the primal and dual variables are updated by
\begin{align}
\bm \theta_f^{(t+1)} \leftarrow & \bm \theta_f^{(t)}+ \frac{\delta_f}{|\mathcal{B}|} \sum_{\mathbf{h}\in \mathcal{B}} \nabla_{\bm \theta_f} \tilde{\mathbf{f}}(\mathbf h; \bm \theta_{f}) \Big[\nabla_{{\mathbf x}}  J(\mathbf x, \mathbf h)   \nonumber \\
& - \nabla_{{\mathbf x}} \mathbf{g}(\mathbf x, \mathbf h) \tilde{\bm \lambda}(\mathbf{h};\bm \theta_\lambda^{(t)} ) - \nabla_{\mathbf{x}} \mathbf c(\mathbf x, \mathbf h) \bm \xi^{(t)}\Big] \Big|_{\mathbf x = \tilde{\mathbf{f}}(\mathbf h; \bm \theta_{f}^{(t)}) }   \nonumber \\
\bm \theta_\lambda^{(t+1)}  \leftarrow& \bm \theta_\lambda^{(t)} + \frac{\delta_\lambda}{|\mathcal{B}|} \sum_{\mathbf{h}\in \mathcal{B}}\nabla_{\bm \theta_\lambda} \tilde{\bm \lambda} (\mathbf h; \bm \theta_\lambda^{(t)}) \mathbf{g}(\tilde{\mathbf{f}}(\mathbf h; \bm \theta_{f}^{(t)} ),\mathbf h)  \nonumber\\
\bm \xi^{(t+1)} \leftarrow&  \bigg[\bm \xi^{(t)} + \frac{\delta_\xi}{|\mathcal{B}|} \sum_{\mathbf{h}\in \mathcal{B}} \mathbf c(\tilde{\mathbf f}(\mathbf h; \bm \theta_f^{(t)}), \mathbf{h})\bigg]^+ \label{eqn:up1}
\end{align}
where $\delta_f$, $\delta_\lambda$ and $\delta_\xi$ are learning rates,  and $[\mathbf x]^+ = [\max(x_1, 0),\cdots, \max(x_2, 0)]^T$. The operation $[\cdot]^+$ in~\eqref{eqn:up1} ensures constraint~\eqref{eqn:c}. Constraints~\eqref{eqn:g} can be satisfied by properly chosen the activation function of the output layer of $\tilde{\bm \lambda}(\mathbf h;\bm \theta_\lambda)$, e.g., \texttt{ReLU}.

If the gradients $\nabla_{{\mathbf x}} J(\mathbf x, \mathbf h)$, $\nabla_{{\mathbf x}}\mathbf g (\mathbf x, \mathbf h)$ and $\nabla_{{\mathbf x}}\mathbf c(\mathbf x, \mathbf h)$ can be computed, an approximated optimal solution of problem $\mathsf{P3}$ can be obtained after the iterations in \eqref{eqn:up1} converges.

\subsection{Model-Free Unsupervised Learning}
For many problems in wireless networks, one or all of the objective and constraint functions in problem $\mathsf{P1}$ cannot be derived in closed-form, and we can only observe the values of these functions after executing $\mathbf x$ at a realization of~$\mathbf h$ and then observe the values of $J(\mathbf x, \mathbf h)$,  $\mathbf g (\mathbf x, \mathbf h)$, and $\mathbf c(\mathbf x, \mathbf h)$. For example, we can measure the data rate $J(\mathbf x, \mathbf h)$ after transmit with power $\mathbf x$ at channel state~$\mathbf h$. For these scenarios, the gradients cannot be derived analytically. In what follows, we resort to model-free unsupervised learning that does not require the explicit expressions of these gradients.

Again according to the universal approximation theorem, we can approximate the objective function and constraints in problem $\mathsf{P1}$ by neural networks as $J(\mathbf x, \mathbf h) \approx \tilde{J}(\mathbf x, \mathbf h;\bm \theta_J)$, $\mathbf g(\mathbf x, \mathbf h) \approx \tilde{\mathbf g}(\mathbf x, \mathbf h; \bm \theta_g)$, and $\mathbf c(\mathbf x, \mathbf h) \approx \tilde{\mathbf c}(\mathbf x, \mathbf h; \bm \theta_c)$. With the approximated objective function and constraints (called as \emph{value networks}), the gradients can be then computed.

The values of $J(\mathbf x, \mathbf h)$, $\mathbf g(\mathbf x, \mathbf h)$, and $\mathbf c(\mathbf x, \mathbf h)$ can be measured and recorded in a system, which can be used as labels for training. Then, the neural networks $\tilde J(\mathbf x, \mathbf h;\bm \theta_J)$, $\tilde{\mathbf g} (\mathbf x, \mathbf h; \bm \theta_g) $, and $\tilde{\mathbf c}(\mathbf x, \mathbf h; \bm \theta_c)$ can be trained by minimizing the $L_2$-norm loss function with stochastic gradient descent as
\begin{align}
\bm \theta_J^{(t+1)} & \leftarrow \bm \theta_J^{(t)} \!- \frac{\delta_J}{|\mathcal{B}|} \!\! \sum_{(\mathbf x, \mathbf h, J) \in \mathcal{B}} \!\!\!\!\! \nabla_{\bm \theta_J} \!\left[ J(\mathbf x, \mathbf h) - \tilde{J}(\mathbf x, \mathbf h;\bm \theta_J^{(t)}) \right]^2  \nonumber  \\
\bm \theta_g^{(t+1)} & \leftarrow \bm \theta_g^{(t)} - \frac{\delta_g}{|\mathcal{B}|}\!\! \sum_{(\mathbf x, \mathbf h, \mathbf g) \in \mathcal{B}}  \!\!\!\!\! \nabla_{\bm \theta_g} \!\left\|  \mathbf g(\mathbf x, \mathbf h) - \tilde{\mathbf{g}}(\mathbf x, \mathbf h;\bm \theta_g^{(t)}) \right\|^2 \nonumber\\
\bm \theta_c^{(t+1)} & \leftarrow \bm \theta_c^{(t)} - \frac{\delta_c}{|\mathcal{B}|} \!\!\sum_{(\mathbf x, \mathbf h, \mathbf c) \in \mathcal{B}}\!\!\!\!\!\! \nabla_{\bm \theta_c} \!\left\|  \mathbf c(\mathbf x, \mathbf h) - \tilde{\mathbf{c}}(\mathbf x, \mathbf h;\bm \theta_c^{(t)}) \right\|^2 \!\!\! \label{eqn:l2}
\end{align}
where $\delta_J$, $\delta_g$ and $\delta_c$ are learning rates, $\mathcal{B}$ denotes a batch of tuples whose elements are the realizations of $\mathbf h$, the corresponding vector $\mathbf x$ conditioned on $\mathbf h$, and the values of $J (\mathbf x, \mathbf h)$, $\mathbf g (\mathbf x, \mathbf h)$ and $\mathbf c (\mathbf x, \mathbf h)$ measured after executing $\mathbf x$.

In the following, we denote $\mathbf y \triangleq \mathbf y (\cdot)$ for notational simplicity, e.g., $J \triangleq J (\mathbf x, \mathbf h)$ and $\tilde J \triangleq \tilde J (\mathbf x, \mathbf h; \bm \theta_J)$. By substituting $\nabla_{\mathbf x} J \approx \nabla_{\mathbf x} \tilde J$, $\nabla_{\mathbf x} \mathbf g \approx \nabla_{\mathbf x} \tilde{\mathbf g}$, and $\nabla_{\mathbf x} \mathbf c \approx \nabla_{\mathbf x} \tilde{\mathbf c}$ into~\eqref{eqn:up1}, we can obtain the update rule for $\bm \theta_f$, $\bm \theta_\lambda$, and $\bm \xi$  as
	\begin{align}
	\bm \theta_f^{(t+1)} &\!\leftarrow  \bm \theta_f^{(t)} \!+\! \frac{\delta_f}{|\mathcal{B}|} \sum_{\mathbf{h}\in \mathcal{B}} \!\! \nabla_{\bm \theta_f} \tilde{\mathbf{f}}\left[\nabla_{{\mathbf x}}  \tilde J  - (\nabla_{{\mathbf x}} \tilde{\mathbf{g}})\tilde{\bm \lambda}   - (\nabla_{\mathbf{x}} \tilde{\mathbf c}) \bm \xi\right] \Big|_{\mathbf x = \tilde{\mathbf{f}} }     \nonumber\\
	\bm \theta_\lambda^{(t+1)} \!& \leftarrow \bm \theta_\lambda^{(t)} + \frac{\delta_\lambda}{|\mathcal{B}|} \sum_{\mathbf{h}\in \mathcal{B}}(\nabla_{\bm \theta_\lambda} \tilde{\bm \lambda}) {\mathbf{g}}\nonumber\\
	\bm \xi^{(t+1)} \!& \leftarrow \bigg[\bm \xi^{(t)} + \frac{\delta_\xi}{|\mathcal{B}|} \sum_{\mathbf{h}\in \mathcal{B}} {\mathbf c}\bigg]^+ \label{eqn:up2}
	\end{align}

\begin{remark}
 When problem $\mathsf{P1}$ has no constraints, our model-free unsupervised learning framework degenerates into a special case of reinforcement learning, where the policy $\mathbf{f}(\mathbf h)$ does not affect the distribution of state~$\mathbf h$. For the unconstrained problem, the gradient of the Lagrangian with respect to policy parameter $\bm \theta_f$ in~\eqref{eqn:pg} degenerates into
\begin{multline}
\nabla_{\bm \theta_f}\mathbb{E}_{\mathbf h} \left[ J( \tilde{\mathbf f}(\mathbf{h};\bm \theta_f), \mathbf h) \right] \\= \mathbb{E}_{\mathbf h}\Big[ \nabla_{\bm \theta_f} \tilde{\mathbf{f}}(\mathbf h; \bm \theta_{f}) \nabla_{{\mathbf x}}  J(\mathbf x, \mathbf h) \big|_{\mathbf{x} = \tilde{\mathbf{f}}(\mathbf h; \bm \theta_{f})} \Big] \label{eqn:dpg}
\end{multline}
which coincides with the  deterministic policy gradient (DPG) theorem~\cite{silver2014deterministic}, where $J(\mathbf x, \mathbf h)$ is actually the action-value function (also known as \emph{Q-function} or \emph{critic}) and the policy network $ \tilde{\mathbf{f}}(\mathbf h; \bm \theta_{f})$ is the \emph{actor}. By replacing $J$  in \eqref{eqn:dpg} with its approximation $\tilde J$, we can obtain the approximated policy gradient used for updating the actor in deep deterministic policy gradient (DDPG) algorithm~\cite{DDPG}.
\end{remark}

Inspired by the great success of \emph{actor-critic} approach in reinforcement learning, we can train the neural networks $\tilde J$, $\tilde{\mathbf g}$, $\tilde{\mathbf c}$, $\tilde{\mathbf f}$, and $\tilde{\bm \lambda}$ simultaneously via interactions with the environment. Each time after we observe the values of $J$, $\mathbf g$, and $\mathbf c$,  we update parameters $\bm \theta_J$, $\bm \theta_g$, and $\bm \theta_c$ to obtain a better approximation of the Lagrangian. Meanwhile, we also update parameters $\bm \theta_f$, $\bm \theta_\lambda$, and $\bm \xi$ to improve the policy. Because $J$, $\mathbf g$, and $\mathbf c$ are functions of $\mathbf x$, to better approximate $J$, $\mathbf g$, $\mathbf c$ and their gradients at $\mathbf x$, it is necessary to obtain the values of $J$, $\mathbf g$, and $\mathbf c$ in the neighborhood of $\mathbf x$. To encourage such exploration, we add a noise term $\mathbf n^{(t)}$ that reduces over iterations to the output of policy network, i.e., $\mathbf x = \tilde {\mathbf f} + \mathbf n^{(t)}$. The detailed learning procedure is provided in Algorithm~\ref{alg1}.

\begin{algorithm}[!htb]
	\caption{\small Model-Free Unsupervised Learning (Deterministic)}\
	\label{alg1}
	\small
	\begin{algorithmic}[1]
		\STATE Initialize neural networks $\tilde J$, $\tilde {\mathbf g}$, $\tilde {\mathbf c}$, $\tilde {\mathbf f}$, $\tilde{\bm \lambda}$ with random parameters $\bm \theta_{J}$, $\bm \theta_{g}$, $\bm \theta_c$, $\bm \theta_f$, $\bm \theta_\lambda$ and initialize multiplier $\bm \xi$.
		\STATE Initialize replay memory $\mathcal{D}$.
		\FOR{ $t = 1, 2, \cdots$}
		\STATE Observe $\mathbf h^{(t)}$ from the environment.
		\STATE Execute $\mathbf x^{(t)} = \tilde{\mathbf f}(\mathbf h^{(t)};\bm \theta_f^{(t)}) + \tilde{\mathbf n}^{(t)}$.
		\STATE Observe values of $J^{(t)} = J(\mathbf x^{(t)}, \mathbf h^{(t)})$, $\mathbf g^{(t)} = \mathbf g(\mathbf x^{(t)}, \mathbf h^{(t)})$, and $\mathbf c^{(t)} = \mathbf c(\mathbf x^{(t)}, \mathbf h^{(t)})$ from the system.
		\STATE Store $\mathbf e^{(t)} = [\mathbf h^{(t)}, \mathbf{x}^{(t)}, J^{(t)}, \mathbf g^{(t)}, \mathbf c^{(t)}]$  in $\mathcal D$.
		\STATE Randomly sample a batch of training samples from~$\mathcal{D}$ as $\mathcal{B}$.
		\STATE Update  $\bm \theta_{J}$, $\bm \theta_{g}$, $\bm \theta_c$ by~\eqref{eqn:l2} and update  $\bm \theta_f$, $\bm \theta_\lambda$, $\bm \xi$ by~\eqref{eqn:up2}.
		\ENDFOR
	\end{algorithmic}
\end{algorithm}

So far, we have implicitly assumed that the policy to be learned is continuous (i.e., $\mathbf{f}(\mathbf{h})$ is a continuous function of~$\mathbf{h}$), and learn its parameterized form $\tilde{ \mathbf{f}}(\mathbf{h} ;\bm \theta_f)$ as a deterministic policy in both model-based and model free unsupervised learning frameworks. In some scenarios, we need to find a discrete policy, e.g., for user scheduling, where parameterizing a deterministic policy is not applicable because the output of neural network is continuous w.r.t the input. Although a discrete policy can be obtained by discretized
a learned deterministic policy, the constraints may not be
satisfied after the discretization.

Alternatively, we can parameterize a stochastic policy by neural network, which can be used to learn both continuous and  discrete policies. Let $\pi(\mathbf x |  \mathbf h; \bm \theta_\pi)$ denote the probability that we execute $\mathbf x$ conditioned on $\mathbf h$, and $\bm \theta_\pi$ is the network parameter. In this case, the parameterized form of problem $\mathsf{P1}$ becomes
\begin{subequations} \label{eqn:p4}
	\begin{align}
	\hspace{-1cm}\mathsf{P4}:\quad	\max_{\bm \theta_\pi} ~& \mathbb{E}_{\mathbf h, \mathbf x \sim \pi } [ J(\mathbf x, \mathbf{h})] \label{eqn:obj4} \\
	\rm{s.t.} ~ & \mathbf g(\mathbf x, \mathbf{h}) \preceq \mathbf{0}  \\
	& \mathbb{E}_{\mathbf h, \mathbf x \sim \pi } [ \mathbf c(\mathbf x, \mathbf{h}) ]\preceq \mathbf{0}
	\end{align}
\end{subequations}
where $\mathbf x \sim \pi$ denotes that random variable $\mathbf x$ is sampled from distribution $\pi(\mathbf x| \mathbf h;\bm \theta_\pi )$, the objective function and average constraints are also averaged over $\mathbf{x}$. We can obtain the Lagrangian and use neural network $\tilde {\bm \lambda}(\mathbf h; \bm \theta_\lambda)$ to parameterize $\bm \lambda (\mathbf h)$. Then, the gradient of Lagrangian with respect to $\bm \theta_\pi$ can be derived as
\begin{subequations}
\begin{align}
&\nabla_{\bm \theta_{\pi}}  \mathcal{L} = \nabla_{\bm \theta_{\pi}}  \mathbb{E}_{\mathbf h,\mathbf x \sim \pi} \left[J- \tilde{\bm \lambda}^T \mathbf g - \bm \xi^T \mathbf c\right] \nonumber \\
&= \mathbb{E}_{\mathbf h} \left[  \sum_{\mathbf x}  \nabla_{\bm \theta_{\pi}} \pi(\mathbf x | \mathbf h; \bm \theta_\pi) (J - \tilde{\bm \lambda}^T \mathbf g - \bm \xi^T \mathbf c) \right] \label{eqn:baseline}\\
& = \mathbb{E}_{\mathbf h} \left[ \sum_{\mathbf x} \pi(\mathbf x | \mathbf h; \bm \theta_\pi ) \frac{\nabla_{\bm \theta_{\pi}} \pi(\mathbf x | \mathbf h; \bm \theta_\pi)}{\pi(\mathbf x | \mathbf h; \bm \theta_\pi )} (J - \tilde{\bm \lambda}^T \mathbf g - \bm \xi^T \mathbf c) \right] \nonumber\\
& = \mathbb{E}_{\mathbf h, \mathbf x \sim \pi}  \left[ (J - \tilde{\bm \lambda}^T \mathbf g - \bm \xi^T \mathbf c) \nabla_{\bm \theta_{\pi}}\log(\pi(\mathbf x | \mathbf h; \bm \theta_\pi )) \right] \label{eqn:policygrad}
\end{align}
\end{subequations}

The gradient of Lagrangian with respect to $\bm \theta_\lambda$ and $\bm \xi$ can be derived as
\begin{align}
&\nabla_{\bm \theta_{\lambda}}  \mathcal{L} =  -  \mathbb{E}_{\mathbf h,\mathbf x \sim \pi} \left[(\nabla_{\bm \theta_\lambda}\tilde{\bm \lambda}) \mathbf g \right] \label{eqn:lambda}\\
& \nabla_{\bm \xi}  \mathcal{L} = -   \mathbb{E}_{\mathbf h,\mathbf x \sim \pi} [ \nabla_{\bm \xi} \mathbf c  ] \label{eqn:xi}
\end{align}

Different from the deterministic policy case, the gradients $\nabla_{\mathbf x} J$, $\nabla_{\mathbf x} \mathbf g$, and $\nabla_{\mathbf x} \mathbf c$ are no longer necessary when we update $\bm \theta_\pi$, $\bm \theta_\lambda$, and $\bm \xi$ with stochastic gradient method. To compute a sample of the gradient in \eqref{eqn:policygrad}$\sim$\eqref{eqn:xi}, we only need to observe the value of $J$, $\mathbf g$, and $\mathbf c$ from the environment\footnote{When model is available, we can compute the values $J$, $\mathbf g$, and $\mathbf c$ from their expressions.} when executing $\mathbf x$ at state $\mathbf h$. Therefore, $\bm \theta_\pi$, $\bm \theta_\lambda$, and $\bm \xi$ are updated by
\begin{subequations} \label{eqn:up3}
\begin{align}
\bm \theta_\pi^{(t+1)} & \leftarrow  \bm \theta_\pi^{(t)} +  \delta_f  (J - \tilde{\bm \lambda}^T \mathbf g - \bm \xi^T \mathbf c) \nabla_{\bm \theta_{\pi}}\log(\pi(\mathbf x | \mathbf h; \bm \theta_\pi )) \label{eqn:stopi}\\
\bm \theta_\lambda^{(t+1)} & \leftarrow \bm \theta_\lambda^{(t)} + \delta_\lambda (\nabla_{\bm \theta_\lambda} \tilde{\bm \lambda}) \mathbf{g}\label{eqn:stol}\\
\bm \xi^{(t+1)} & \leftarrow \left[\bm \xi^{(t)} + \delta_\xi  \mathbf c\right]^+ \label{eqn:stoxi}
\end{align}
\end{subequations}

\begin{remark}
Although~\eqref{eqn:policygrad} is derived assuming discrete distribution of $\mathbf{x}$, it can also be derived from a continuous distribution of $\mathbf x$. Therefore, \eqref{eqn:up3} (and the following updating rules in~\eqref{eqn:policygrad2}) are also applicable for learning a continuous policy.
\end{remark}
\begin{remark}
	When there are no constraints in problem~$\mathsf{P4}$, \eqref{eqn:policygrad} reduces to
	\begin{multline}
	\nabla_{\bm \theta_{f}} \mathbb{E}_{\mathbf h, \mathbf x \sim \pi} \left[ J(\mathbf x, \mathbf h) \right] \\
	= \mathbb{E}_{\mathbf h, \mathbf x \sim \pi}  \left[ J(\mathbf x, \mathbf h) \nabla\log(\pi(\mathbf x | \mathbf h; \bm \theta_\pi )) \right]
	\end{multline}
	which coincides with the policy gradient theorem~\cite{sutton1998reinforcement} in reinforcement learning, and the update of $\bm \theta_\pi$ in~\eqref{eqn:up3} degenerates into the REINFORCE method~\cite{sutton1998reinforcement}.
\end{remark}

The stochastic gradient update in~\eqref{eqn:stopi} may exhibit large variance~\cite{sutton1998reinforcement} because the parameterized policy is stochastic and hence converge slowly. Inspired by the advantage actor-critic approach~\cite{A3C}, we can subtract a term $\mathbb{E}_{\mathbf x\sim \pi} [ J(\mathbf x, \mathbf h) - \tilde{\bm \lambda}^T \mathbf g (\mathbf x, \mathbf h)- \bm \xi^T \mathbf c (\mathbf x, \mathbf h)]$
into the parenthesis of~\eqref{eqn:baseline}, which do not change the expectation of gradients but can reduce the variance. Then, the update for $\bm \theta_\pi$ becomes
\begin{multline}
\bm \theta_{\pi}^{(t + 1)} \leftarrow \bm \theta_{\pi}^{(t)}+  \delta_\pi \Big[(J - \mathbb{E}_{\mathbf x \sim \pi} [ J])  - \tilde{\bm \lambda}^T (\mathbf g - \mathbb{E}_{\mathbf x \sim \pi} [\mathbf g]) \\
- \bm \xi^T (\mathbf c - \mathbb{E}_{\mathbf x \sim \pi}[\mathbf c]) \Big]  \nabla_{\bm \theta_\pi}\log(\pi(\mathbf x | \mathbf h; \bm \theta_\pi^{(t)}) \label{eqn:policygrad2}
\end{multline}

Again, the average terms can be approximated by neural networks as $\mathbb{E}_{\mathbf x \sim \pi} [J] \approx \bar J (\mathbf h; \bm \theta_{\bar J})$, $\mathbb{E}_{\mathbf x \sim \pi} [\mathbf g] \approx \bar{\mathbf g} (\mathbf h; \bm \theta_{\bar g})$ and $\mathbb{E}_{\mathbf x \sim \pi} [\mathbf c] \approx \bar{\mathbf c} (\mathbf h; \bm \theta_{\bar c})$, which are updated by minimizing the $L_2$-norm loss with stochastic gradient descent as
\begin{align}
\bm \theta_{\bar J}^{(t+1)} & \leftarrow \bm \theta_{\bar J}^{(t)} - \delta_{\bar J}\nabla_{\bm \theta_{\bar J}} \left[ J(\mathbf x, \mathbf h) - \bar{J}(\mathbf h;\bm \theta_{\bar J}^{(t)}) \right]^2  \nonumber  \\
\bm \theta_{\bar g}^{(t+1)} & \leftarrow \bm \theta_{\bar g}^{(t)} - \delta_{\bar g}\nabla_{\bm \theta_{\bar g}} \left\|  \mathbf g(\mathbf x, \mathbf h) - \bar{\mathbf{g}}(\mathbf h;\bm \theta_{\bar g}^{(t)}) \right\|^2 \nonumber\\
\bm \theta_{\bar c}^{(t+1)} & \leftarrow \bm \theta_{\bar c}^{(t)} - \delta_{\bar c}  \nabla_{\bm \theta_{\bar c}} \left\|  \mathbf c(\mathbf x, \mathbf h) - \bar {\mathbf{c}}(\mathbf h;\bm \theta_{\bar c}^{(t)}) \right\|^2  \label{eqn:l22}
\end{align}

The detailed learning procedure is provided in Algorithm~\ref{alg2}.
\begin{algorithm}[!htb]
	\caption{\small Model-Free Unsupervised Learning (Stochastic)}\
	\label{alg2}
	\small
	\begin{algorithmic}[1]
		\STATE Initialize neural networks $\bar J$, $\bar {\mathbf g}$, $\bar {\mathbf c}$, $\pi$, $\tilde{\bm \lambda}$ with random parameters $\bm \theta_{\bar J}$, $\bm \theta_{\bar g}$, $\bm \theta_c$, $\bm \theta_\pi$, $\bm \theta_\lambda$ and initialize multiplier $\bm \xi$.
		\FOR{ $t = 1, 2, \cdots$}
		\STATE Observe $\mathbf h^{(t)}$ from the environment.
		\STATE Sample $\mathbf x^{(t)}$ from $\pi (\mathbf{x}^{(t)} | \mathbf h^{(t)}; \bm \theta_\pi^{(t)})$ and execute $\mathbf x^{(t)}$.
		\STATE Observe values of $J^{(t)} = J(\mathbf x^{(t)}, \mathbf h^{(t)})$, $\mathbf g^{(t)} = \mathbf g(\mathbf x^{(t)}, \mathbf h^{(t)})$, and $\mathbf c^{(t)} = \mathbf c(\mathbf x^{(t)}, \mathbf h^{(t)})$.
		\STATE Update  $\bm \theta_{\bar J}$, $\bm \theta_{\bar g}$, $\bm \theta_{\bar c}$ by~\eqref{eqn:l22}, update  $\bm \theta_\pi$ by substituting $\mathbb{E}_{\mathbf x \sim \pi} [J] \approx \bar J (\mathbf h^{(t)}; \bm \theta_{\bar J}^{(t)})$, $\mathbb{E}_{\mathbf x \sim \pi} [\mathbf g] \approx \bar{\mathbf g} (\mathbf h^{(t)}; \bm \theta_{\bar g}^{(t)})$ and $\mathbb{E}_{\mathbf x \sim \pi} [\mathbf c] \approx \bar{\mathbf c} (\mathbf h^{(t)}; \bm \theta_{\bar c}^{(t)})$ into~\eqref{eqn:policygrad2}, and update $\bm \theta_\lambda$ and $\bm \xi$ by \eqref{eqn:stol} and \eqref{eqn:stoxi}, repsectively.
		\ENDFOR
	\end{algorithmic}
\end{algorithm}

\section{Case Study: Power Control Problem}
In this section, we illustrate how to apply the model-based and model-free unsupervised learning frameworks for solving optimization problems. For easy understanding, we consider a simple power control problem in point to point communications. To provide a baseline, we first derive the analytical solution of the problem. Then, we show how to employ the frameworks when the expression of the objective function is known and unknown.

In what follows we optimize the instantaneous transmit power to minimize the ergodic capacity under the constraints of average transmit power and maximum transmit power,
\begin{align}\label{prob:PwrCtrl}
\hspace{-1cm}\mathsf{P5}:~	\max \limits_{P(h)} \quad & \mathbb{E}_h \left[R\left(P(h),h\right)\right] \\
\rm{s.t.} ~ & \mathbb{E}_h \left[P(h)\right] \leq \bar{P} \label{con:Pmean} \tag{\theequation a} \\
& 0 \leq P(h) \leq P_{\max}, \  \forall h \label{con:Pmax} \tag{\theequation b}
\end{align}
where $h$ is the small-scale channel gain, $P(h)$ is the transmit power adapted to $h$, $R(P(h),h)$ is the channel capacity, $\bar{P} >0$ is the maximum average transmit power, and $P_{\max} > \bar{P}$ is the maximum instantaneous transmit power.

\subsection{Analytical Solution}
When the channel coding is sufficient long and the noise is Gaussian distributed, the channel capacity can be expressed as the Shannon's formula, i.e., $R\left(P(h),h\right) \!=\! \log_2 \!{(1 \!+\! \frac{h P(h)}{N})}$, where $N \!>\! 0$ is the power of noise unified by the large-scale channel gain. Then the Karush-Kuhn-Tucker (KKT) conditions of problem~\eqref{prob:PwrCtrl} can be derived as~\cite{gregory2018constrained},
\begin{subequations}\label{KKT}
	\begin{align}
	\frac{1}{N/h + P(h)} + \lambda_1(h) - \lambda_2(h) - \xi &= 0 \label{KKT:PwrCtrl} \\
	\xi \left(\mathbb{E}_h \left[P(h)\right] - \bar{P}\right) &= 0 \label{KKT:Mean} \\
	\lambda_1(h) P(h) &= 0, \  \forall h \label{KKT:Min} \\
	\lambda_2(h) \left(P(h) - P_{\max}\right) &= 0, \  \forall h \label{KKT:Max} \\
	\eqref{con:Pmean}, \ ~\eqref{con:Pmax}, \  \xi \geq 0, \  \lambda_1(h), \lambda_2(h) &\geq 0, \  \forall h
	\end{align}
\end{subequations}
As proved in the Appendix, the solution of the problem is,
\begin{align}\label{eq:WaterFilling}
P^*(h) = \left\{
\begin{array}{ll}
0, & h \leq \xi^* N \\
1/\xi^* - N/h, & \xi^* N < h < \frac{N}{1/\xi^* - P_{\max}} \\
P_{\max}, & h \geq \frac{N}{1/\xi^* - P_{\max}}
\end{array}
\right.
\end{align}
where $\xi^*$ satisfies $\mathbb{E}_h \left[P^*(h)\right] = \bar{P}$ and can be computed via bisection searching with known distribution of $h$. The solution in~\eqref{eq:WaterFilling} differs from the water-filling structure~\cite{goldsmith2005wireless} due to the additional constraint imposed by $P_{\max}$.

\vspace{-1mm}
\subsection{Model-Based Unsupervised Learning Method}
Problem $\mathsf{P5}$ may not have closed-form solution, say when the finite block-length channel coding is used such that $R\left(P(h),h\right)$ is with complex expression. In the sequel, we illustrate how to use model-based unsupervised learning to solve the problem.

The function to be optimized is approximated by a policy network $\tilde{P}(h;\bm \theta_P)$. The constraints in~\eqref{con:Pmax} can be satisfied by setting the active function of the output layer in $\tilde{P}(h;\bm \theta_P)$ as \texttt{Sigmoid}, and multiplying the final output by $P_{\max}$. However, to validate the effectiveness of the multiplier network in handling the instantaneous constraints in functional optimization problems, we use \texttt{ReLU} as the active function of the output layer to only ensure $\tilde{P}(h; \bm \theta_P) \geq 0$, and introduce the multiplier network $\tilde{\lambda}(h;\bm \theta_\lambda)$ to ensure the constraint $\tilde{P}(h;\bm \theta_P) \leq P_{\max}$ with primal-dual stochastic gradient method given by~\eqref{eqn:up2}. Then, the power control policy and the Lagrange multipliers can be updated by
\begin{subequations}
\begin{align}
\bm \theta_P^{(t+1)} &\leftarrow  \bm \theta_P^{(t)}+ \frac{\delta_P}{|\mathcal{B}|} \sum_{h\in \mathcal{B}} {\nabla_{\bm \theta_P} \tilde{P}(\nabla_P  R - \tilde{\lambda} - \xi)} \label{eqn:policygradP} \\
\bm \theta_\lambda^{(t+1)} & \leftarrow  \bm \theta_\lambda^{(t)} + \frac{\delta_\lambda}{|\mathcal{B}|} \sum_{h \in \mathcal{B}} {\nabla_{\bm \theta_\lambda} \tilde{\lambda} (\tilde{P} - P_{\max})} \label{eqn:lambdagrad} \\
\xi^{(t+1)} & \leftarrow \bigg[\xi^{(t)} - \frac{\delta_\xi}{|\mathcal{B}|} \sum_{h \in \mathcal{B}} {\tilde{P}}\bigg]^+ \label{eqn:xigrad}
\end{align}
\end{subequations}
where $\delta_P$, $\delta_\lambda$, and $\delta_\xi$ are the learning rates, and $\mathcal{B}$ denotes a batch of training samples.

\subsection{Model-Free Unsupervised Learning Method}
When the channel coding is short or the noise is not Gaussian distributed, the Shannon's formula is not applicable and the expression of $R\left(P(h),h\right)$ is hard to obtain. In the following, we illustrate how to use the proposed model-free unsupervised learning to solve problem $\mathsf{P5}$.

The objective function is approximated by introducing the value network $\tilde{R}(P,h;\bm \theta_R)$, which is then used to compute the approximated gradient $\nabla_P \tilde{R} \approx \nabla_P R$ for updating the policy network parameter $\bm \theta_P$ in~\eqref{eqn:policygradP}. The updates for $\bm \theta_\lambda$ and $\xi$ are the same as in~\eqref{eqn:lambdagrad} and~\eqref{eqn:xigrad} since the expressions of constraints are known. After observing the actual data rate acheived by transmiting with power $P$ at channel state $h$, the value network $\tilde{R}(P,h;\bm \theta_R)$ is trained based on the observed value of $R(P,h)$ according to~\eqref{eqn:l2} as
\begin{equation}\label{trn:R}
\bm \theta_R^{(t+1)}\!  \!\leftarrow\! \bm \theta_R^{(t)} \!- \frac{\delta_R}{|\mathcal{B}|} \!\! \sum_{(P, h, R) \in \mathcal{B}} \!\!\!\!\! \!\!\nabla_{\bm \theta_R} \!\!\left[ R(P,h) - \tilde{R}(P,h;\bm \theta_R) \right]^2 \!
\end{equation}
where  $\delta_R$ denotes the learning rate.

\section{Numerical and Simulation Results}
In this section, we validate the proposed mode-free unsupervised learning frameworks by considering problem $\mathsf{P5}$, where in simulation the channel coding is assumed long and the noise is assumed Gaussian.

The simulation setup is as follows. The maximal instantaneous and average transmit powers are $P_{\max} = 40$ W and  $\bar P = 30$ W, respectively. The distance between the transmitter and the receiver is $d = 500$ m. The noise power spectral density is $-174$ dBm/Hz and the bandwidth is $20$ MHz. We consider Rayleigh fading channels and the path loss is modeled by $35.3 + 37.6 \log_{10}(d)$ in~dB.

The hyper-parameters used for model-based and model-free frameworks are as follows. Both $\tilde \lambda$ and $\tilde P$ have three fully-connected hidden layers with $50$, $40$, and $30$ nodes, respectively. $\tilde R$ has two hidden layers with $200$ and $150$ nodes, respectively. All the hidden layers and the output layers of $\tilde \lambda$ and $\tilde P$ use \texttt{ReLU} as the activation function. The output layer of $\tilde R$ has no activation function. We use Adam~\cite{adam} for training all the neural networks with learning rate $\delta_P = \delta_\lambda = 10^{-3}$ for $\tilde P$ and $\tilde \lambda$, and $\delta_R = 5\times 10^{-3}$ for $\tilde R$. The batch size is $|\mathcal{B}| = 32$. $\tilde P$ is initialized as $10$. Both $\tilde\lambda$ and $\xi$  are initialized as~$0$.  The noise term for exploration in model-free learning is set as $n^{(t)} = \epsilon^{(t)} \mathcal{N}^{(t)}$ where $\mathcal{N}^{(t)}$ denotes Gaussian noise with zero mean and unit variance. The value of $\epsilon^{(t)}$ is set as $10$ for the first $5\times 10^3$ iterations and then decreases linearly to zero for the next $1.5 \times 10^4$ iterations. All the simulation results are averaged over $50$ rounds of learning. 

\begin{figure}[htb]
	\centering	
	\vspace{-2mm}
	\subfigure[Average rate.]{
		\label{fig:rate} 
		\includegraphics[width=0.38\textwidth]{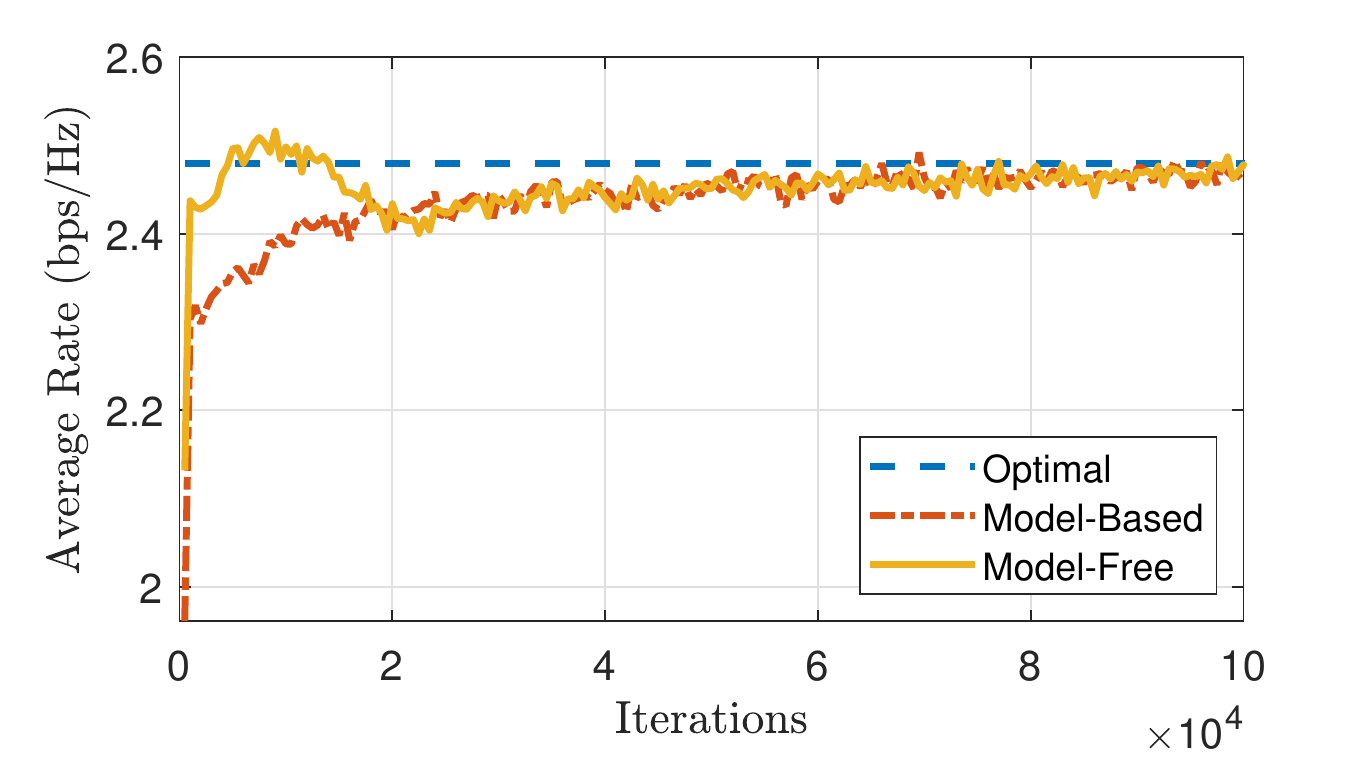}}
	\vspace{-2mm}
	\subfigure[Instanstaneous constraint violation.]{
		\label{fig:in} 
 \includegraphics[width=0.38\textwidth]{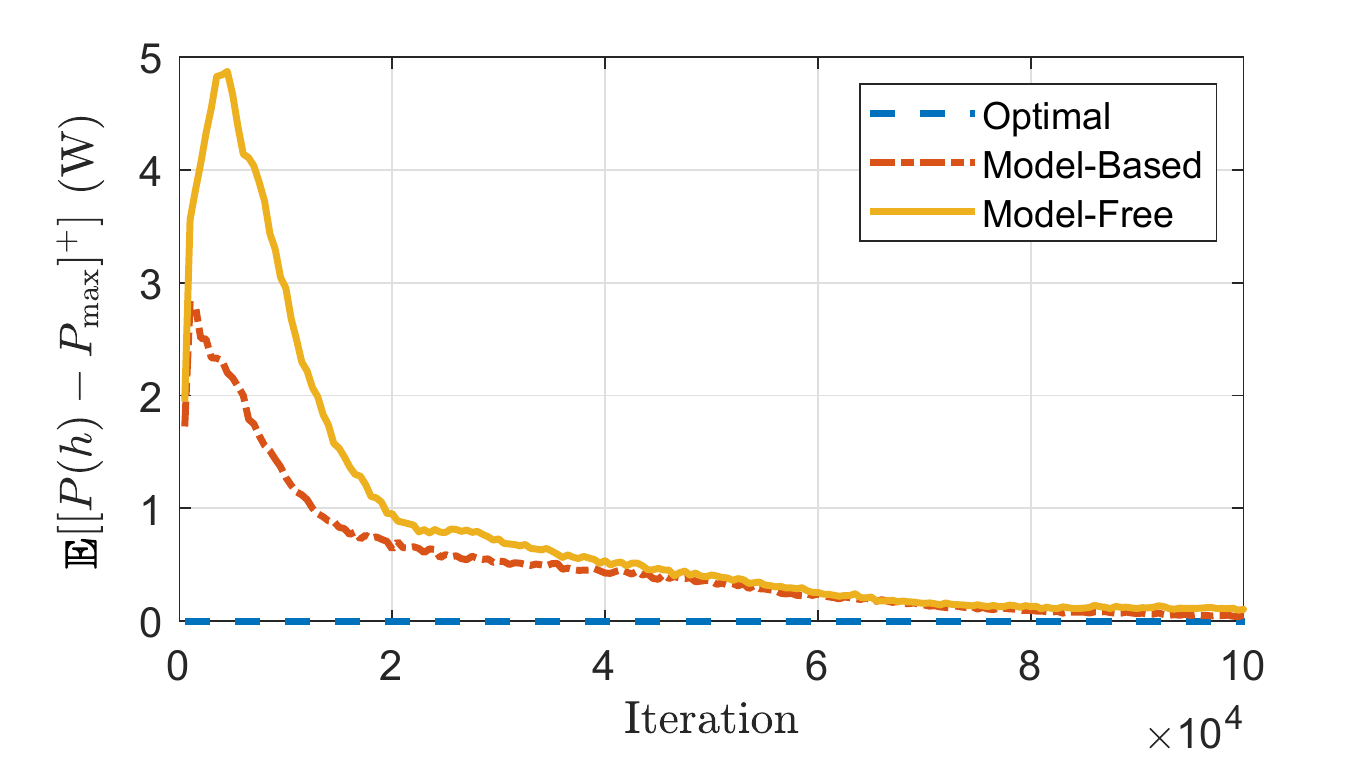}}
	\vspace{-2mm}
	\subfigure[Average constraint violation.]{
		\label{fig:ave} 
		\includegraphics[width=0.38\textwidth]{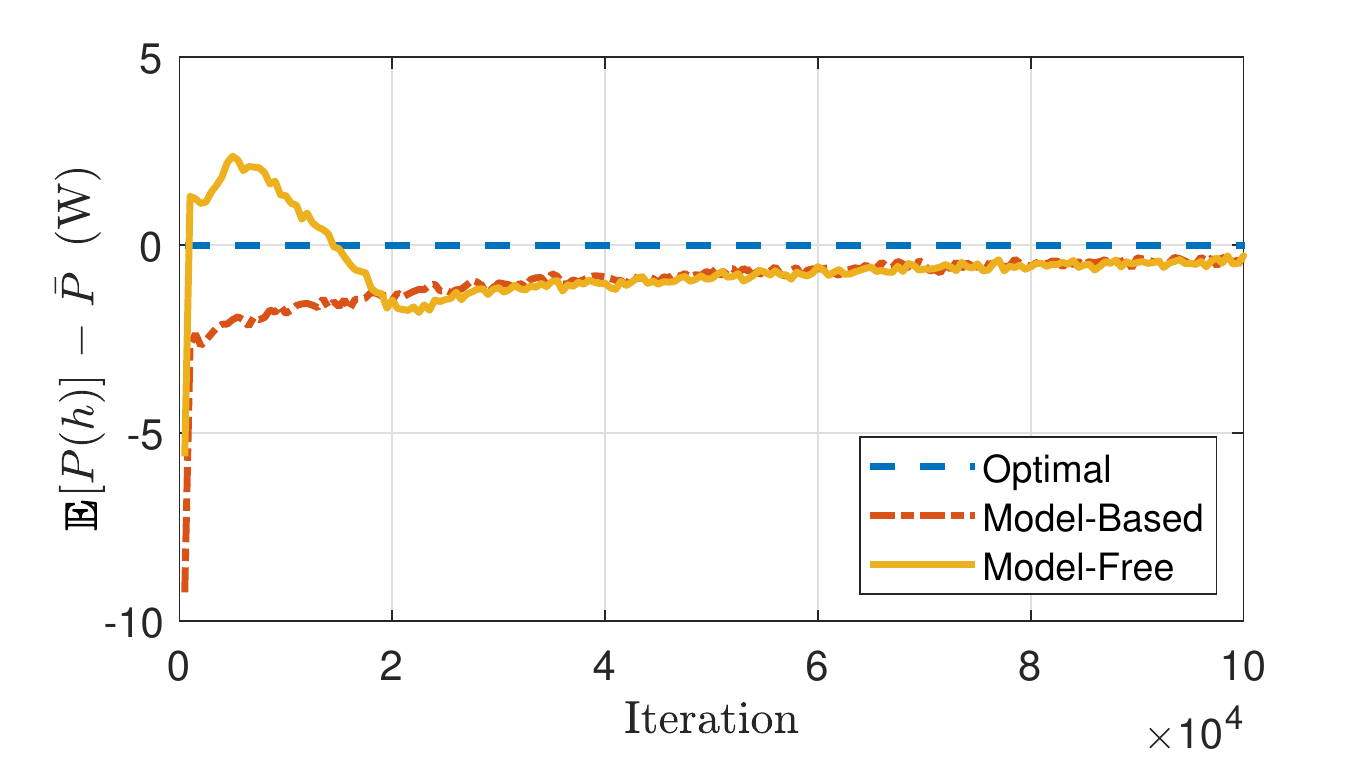}}
	\caption{Convergence comparison. The results are averaged over $500$ successive iterations.}
	\label{fig:converge}
\end{figure}

In Fig.~\ref{fig:converge}, we compare the convergence of model-based and model-free unsupervised learning. Both model-based and model-free learning can converge to the average rate achieved by the optimal solution $P^*(h)$ numerically computed with \eqref{eq:WaterFilling} (with legend ``Optimal"). The average rate achieved by model-free learning can be even higher than the optimal solution at the beginning due to violation of constraints. We show the violations of instantaneous and average constraints in Fig.~\ref{fig:in} and Fig.~\ref{fig:ave}, respectively. Since model-free learning needs the exploration to learn the expression of the objective function, the violations of constraints are more severe than model-based method at the beginning of learning due to insufficient training samples. With the increase of iterations, both model-based and model-free learning can satisfy all the constraints after convergence. Moreover, the number of iterations for converging to the optimal solution are close for model-based and model-free learning. This demonstrates the efficiency of proposed model-free unsupervised learning framework where the policy, multiplier, and value networks are trained simultaneously.
\begin{figure}[!htb]
	\vspace{-3mm}
	\centering
	\includegraphics[width=0.38\textwidth]{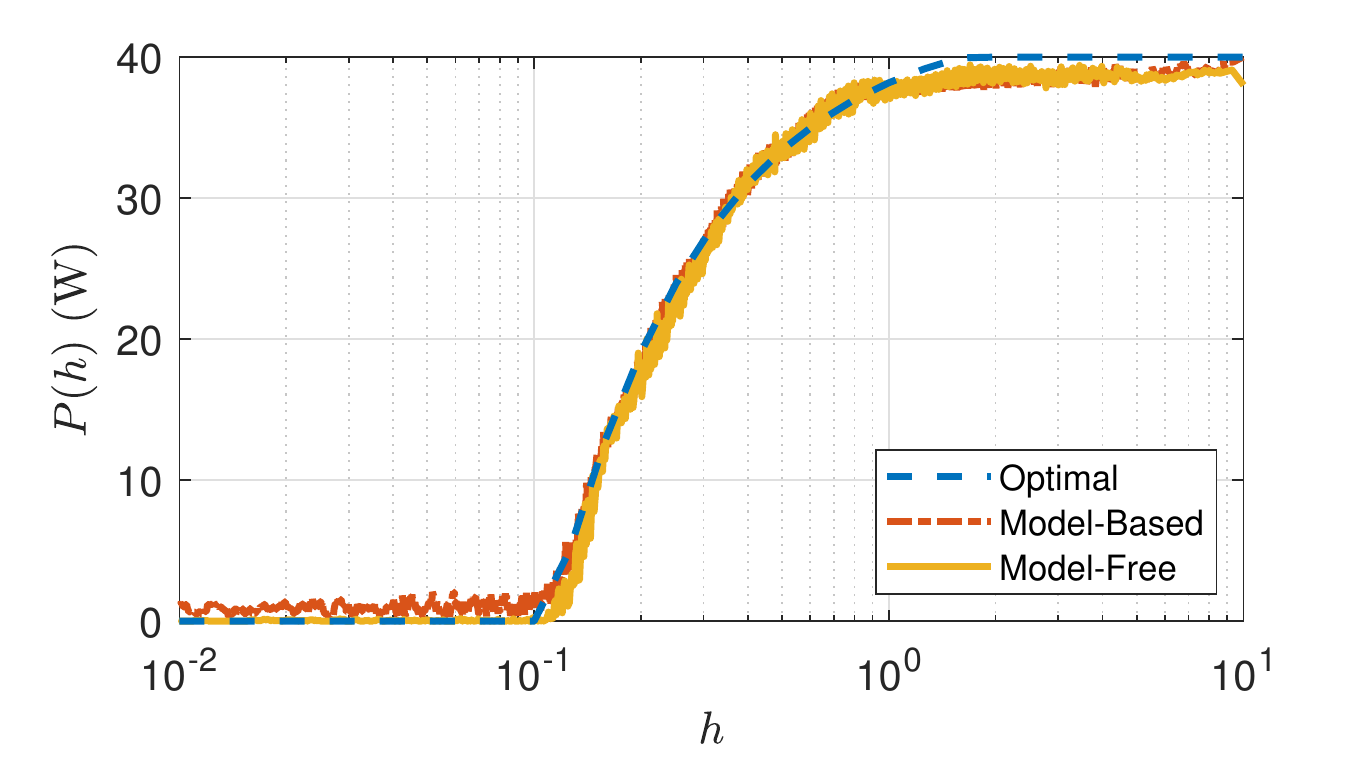}
	\vspace{-3mm}
	\caption{Comparison of learned policy after $10^5$ iterations. }
	\label{fig:policy}
	\vspace{-2mm}
\end{figure}

In Fig.~\ref{fig:policy}, we compare the behavior of the policies learned by model-based and model-free frameworks with the optimal solution. We can see that the learned policies behave almost the same with optimal policy.

\section{Conclusions}
In this paper, we proposed an framework to solve optimization problems with constraints by model-free unsupervised learning, and revealed the connections with reinforcement learning. We illustrated how to apply the proposed framework by a power control problem. Numerical and simulation results validated our framework and showed that model-free unsupervised learning can converge to the optimal policy with similar speed as model-based unsupervised learning.

\appendix

To find the solution from the KKT conditions in \eqref{KKT}, $P^*(h)$, $\lambda_1^*(h)$, $\lambda_2^*(h)$, and $\xi^*$, we first prove $\xi^* > 0$.

Assume $\xi^* = 0$. Since $\frac{1}{N/h + P^*(h)} > 0$ and $\lambda_1^*(h) \geq 0$, we have $\lambda_2^*(h) > 0$ according to \eqref{KKT:PwrCtrl}. Then, $P^*(h) = P_{\max}$ can be derived from \eqref{KKT:Max}. In this case, $\mathbb{E}_h \left[P(h)\right] = P_{\max} > \bar{P}$, which violate the constraint in \eqref{con:Pmean}. Therefore, $\xi^* > 0$. From \eqref{KKT:Mean}, we further have,
\begin{align}\label{eq:Pmean}
  \mathbb{E}_h \left[P^*(h) \right] = \bar{P}
\end{align}

When $h < \xi^* N$, we have $\frac{1}{N/h + P^*(h)} - \xi^* < 0$. In this case, $\lambda_1^*(h) > 0$ according to~\eqref{KKT:PwrCtrl} and $P^*(h) = 0$ according to~\eqref{KKT:Min}.
When $h = \xi^* N$, $\frac{1}{N/h + P^*(h)} - \xi^* < 0$ if $P^*(h) >$$~0$, which on the contrary results in $P^*(h) = 0$. Therefore, $P^*(\xi^* N) = 0$.

When $h > \frac{N}{1/\xi^* -  P_{\max}}$, we have $\frac{1}{N/h + P^*(h)} - \xi^* > 0$. In this case, $\lambda_2^*(h) > 0$ according to \eqref{KKT:PwrCtrl} and $P^*(h) = P_{\max}$ according to \eqref{KKT:Max}.
When $h = \frac{N}{1/\xi^* - P_{\max}}$, $\frac{1}{N/h + P^*(h)} - \xi^* > 0$ if $P^*(h) < P_{\max}$, which results in $P^*(h) = P_{\max}$, contradicting with $P^*(h) < P_{\max}$. Therefore, $P^*(\frac{N}{1/\xi^* - P_{\max}}) = P_{\max}$.

When $\xi^* N < h < \frac{N}{1/\xi^* - P_{\max}}$, we have $\frac{1}{N/h + P_{\max}} < \xi^* < \frac{1}{N/h + 0}$. In this case, if $P^*(h) = 0$, then $\lambda_2^*(h) > 0$ according to \eqref{KKT:PwrCtrl}, which results in $P^*(h) = P_{\max}$, contradicting with $P^*(h) = 0$. Similarly, if $P^*(h) = P_{\max}$, then $\lambda_1^*(h) > 0$ according to \eqref{KKT:PwrCtrl}, which results in $P^*(h) = 0$, contradicting with $P^*(h) = P_{\max}$. Therefore, we have $0 < P^*(h) < P_{\max}$ and $\lambda_1^*(h), \lambda_2^*(h) = 0$. According to \eqref{KKT:PwrCtrl} we further have $P^*(h) = 1/\xi^* - N/h$.

Finally, according to the solution of $P^*(h)$, $\xi^*$ can be solved from \eqref{eq:Pmean}.

\bibliographystyle{IEEEtran}
\bibliography{ref}

\begin{thebibliography}{10}
\providecommand{\url}[1]{#1}
\csname url@samestyle\endcsname
\providecommand{\newblock}{\relax}
\providecommand{\bibinfo}[2]{#2}
\providecommand{\BIBentrySTDinterwordspacing}{\spaceskip=0pt\relax}
\providecommand{\BIBentryALTinterwordstretchfactor}{4}
\providecommand{\BIBentryALTinterwordspacing}{\spaceskip=\fontdimen2\font plus
\BIBentryALTinterwordstretchfactor\fontdimen3\font minus
  \fontdimen4\font\relax}
\providecommand{\BIBforeignlanguage}[2]{{%
\expandafter\ifx\csname l@#1\endcsname\relax
\typeout{** WARNING: IEEEtran.bst: No hyphenation pattern has been}%
\typeout{** loaded for the language `#1'. Using the pattern for}%
\typeout{** the default language instead.}%
\else
\language=\csname l@#1\endcsname
\fi
#2}}
\providecommand{\BIBdecl}{\relax}
\BIBdecl

\bibitem{kong2018hybrid}
L.~Kong, S.~Han, and C.~Yang, ``Hybrid precoding with rate and coverage
  constraints for wideband massive {MIMO} systems,'' \emph{IEEE Trans. Wireless
  Commun.}, vol.~17, no.~7, pp. 4634--4647, Jul. 2018.

\bibitem{liu2018caching}
D.~Liu and C.~Yang, ``Caching at base stations with heterogeneous user demands
  and spatial locality,'' \emph{IEEE Trans. Commun.}, vol.~67, no.~2, pp.
  1554--1569, Feb 2018.

\bibitem{Zeidler2013Functional}
E.~Zeidler, \emph{Nonlinear functional analysis and its applications: III:
  variational methods and optimization}.\hskip 1em plus 0.5em minus 0.4em\relax
  Springer Science \& Business Media, 2013.

\bibitem{goldsmith2005wireless}
A.~Goldsmith, \emph{Wireless Communications}.\hskip 1em plus 0.5em minus
  0.4em\relax Cambridge Univ. Press, 2005.

\bibitem{boyd2004convex}
S.~Boyd and L.~Vandenberghe, \emph{Convex optimization}.\hskip 1em plus 0.5em
  minus 0.4em\relax Cambridge Univ. Press, 2004.

\bibitem{Zienkiewicz1977FEM}
O.~C. Zienkiewicz, R.~L. Taylor, P.~Nithiarasu, and J.~Zhu, \emph{The finite
  element method}.\hskip 1em plus 0.5em minus 0.4em\relax McGraw-hill London,
  1977, vol.~3.

\bibitem{Chengjian2019GC}
C.~Sun and C.~Yang, ``Unsupervised deep learning for ultra-reliable and
  low-latency communications,'' in \emph{Proc. IEEE Globecom, to appear}, 2019.

\bibitem{gregory2018constrained}
J.~Gregory, \emph{Constrained optimization in the calculus of variations and
  optimal control theory}.\hskip 1em plus 0.5em minus 0.4em\relax Chapman and
  Hall/CRC, 2018.

\bibitem{Hornik1989UnivApprox}
K.~Hornik, M.~B. Stinchcombe, and H.~White, ``Multilayer feedforward networks
  are universal approximators,'' \emph{Neural Networks}, vol.~2, no.~5, pp.
  359--366, 1989.

\bibitem{Chengjian2019PIMRC}
C.~Sun and C.~Yang, ``Learning to optimize with unsupervised learning: Training
  deep neural networks for {URLLC},'' in \emph{Proc. IEEE PIMRC, to appear},
  2019.

\bibitem{silver2014deterministic}
D.~Silver, G.~Lever, N.~Heess, T.~Degris, D.~Wierstra, and M.~Riedmiller,
  ``Deterministic policy gradient algorithms,'' in \emph{ICML}, 2014.

\bibitem{DDPG}
T.~P. Lillicrap, J.~J. Hunt, A.~Pritzel, N.~Heess, T.~Erez, Y.~Tassa,
  D.~Silver, and D.~Wierstra, ``Continuous control with deep reinforcement
  learning,'' in \emph{Proc. ICLR}, 2016.

\bibitem{sutton1998reinforcement}
R.~S. Sutton and A.~G. Barto, \emph{Reinforcement learning: An
  introduction}.\hskip 1em plus 0.5em minus 0.4em\relax MIT Press Cambridge,
  1998.

\bibitem{A3C}
V.~Mnih, A.~P. Badia, M.~Mirza, A.~Graves, T.~Lillicrap, T.~Harley, D.~Silver,
  and K.~Kavukcuoglu, ``Asynchronous methods for deep reinforcement learning,''
  in \emph{Proc. ICML}, 2016.

\bibitem{adam}
D.~P. Kingma and J.~Ba, ``Adam: A method for stochastic optimization,'' in
  \emph{Proc. ICLR}, 2014.

\end{thebibliography}

\end{document}